\pdfoutput=1

\documentclass[11pt]{article}

\usepackage[]{emnlp2021}
\usepackage{times}
\usepackage{latexsym}

\usepackage[T1]{fontenc}

\usepackage[utf8]{inputenc}

\usepackage{microtype}
\usepackage{tabularx}
\usepackage{multirow}
\usepackage{booktabs}
\usepackage{graphicx}
\usepackage{ragged2e}
\usepackage{footnote}
\usepackage{bbm}
\usepackage{bm}
\usepackage{latexsym}
\usepackage{amsmath}
\usepackage{amssymb}
\title{Finding a Balanced Degree of Automation for Summary Evaluation}

\author{Shiyue Zhang $\;\;\;\;$ Mohit Bansal \\
  UNC Chapel Hill  \\
  {\tt \{shiyue, mbansal\}@cs.unc.edu} 
}

\begin{document}
\maketitle
\begin{abstract}
Human evaluation for summarization tasks is reliable but brings in issues of reproducibility and high costs. Automatic metrics are cheap and reproducible but sometimes poorly correlated with human judgment. 
In this work, we propose flexible semi-automatic to automatic summary evaluation metrics, following the \emph{Pyramid} human evaluation method. 
Semi-automatic \emph{Lite$^2$Pyramid} retains the reusable human-labeled Summary Content Units (SCUs) for reference(s) but replaces the manual work of judging SCUs' presence in system summaries with a natural language inference (NLI) model. 
Fully automatic \emph{Lite$^3$Pyramid} further substitutes SCUs with automatically extracted Semantic Triplet Units (STUs) via a semantic role labeling (SRL) model. Finally, we propose in-between metrics,
\emph{Lite$^{2.x}$Pyramid}, where we use a simple regressor to predict how well the STUs can simulate SCUs and retain SCUs that are more difficult to simulate, which provides a smooth transition and balance between automation and manual evaluation. 
Comparing to 15 existing metrics, we evaluate human-metric correlations on 3 existing meta-evaluation datasets and our newly-collected \emph{PyrXSum} (with 100/10 XSum examples/systems). 
It shows that \emph{Lite$^2$Pyramid} consistently has the best summary-level correlations; \emph{Lite$^3$Pyramid} works better than or comparable to other automatic metrics; \emph{Lite$^{2.x}$Pyramid} trades off small correlation drops for larger manual effort reduction, which can reduce costs for future data collection.\footnote{Our code and data are publicly available at: \url{https://github.com/ZhangShiyue/Lite2-3Pyramid}}

\end{abstract}

\section{Introduction}
\label{sec:intro}
Evaluating the quality of summaries is a challenging task. Human evaluation is usually regarded as the gold standard. 
Out of different human evaluation methods, \emph{Pyramid} \cite{nenkova-passonneau-2004-evaluating} has been perceived as an objective and reliable protocol and used by early summarization benchmarks, e.g., TAC \cite{DBLP:conf/tac/2008, DBLP:conf/tac/2009}. Given one or several reference summaries of an example, human assessors first exhaustively extract Summary Content Units (SCUs), each SCU contains a single fact, from the reference(s), and then check whether they are present in a system summary. Figure~\ref{fig:metric} shows an example of human-labeled SCUs. 
Despite the reliability, manual evaluation is usually: (1) \emph{not reproducible}, results may change when different evaluators are involved, making it hard to compare the results across papers; (2) \emph{expensive}, in terms of time and cost. Thus, it is unlikely to apply human evaluation extensively in model selection (e.g., to choose the best checkpoint); instead, people usually treat it as an additional quality verification step. Aiming to work as a proxy of humans, many automatic metrics have been proposed \cite{lin2004rouge, tratz2008bewte, giannakopoulos2011autosummeng, yang2016peak, zhang2019bertscore, deutsch2020towards}. However, most of them cannot reliably substitute human evaluation due to the unstable performance across datasets \cite{bhandari-etal-2020-evaluating}, weak to moderate correlations with human judgment \cite{fabbri2021summeval}, or more indication of topic similarity than information overlap \cite{deutsch2020understanding}. 

In this work, we want to combine human and automatic evaluations and find a balance between reliability and reproducibility (plus expense).
Recall the Pyramid method~\cite{nenkova-passonneau-2004-evaluating}, where these SCUs for reference summaries only need to be annotated once, then they can be fixed. It means SCUs can come with the datasets and are reusable for evaluating different systems. Hence, what hinders 
this method from being reproducible is its second step of asking humans to judge the presence of SCUs in system summaries. Whenever we have a new summarizer, we need to collect human labels for this step. Therefore, we propose to retain the reusable SCUs but replace human effort in the second step with a neural model. Basically, people are answering \emph{whether a SCU is entailed by the summary}, which is closely related to the Natural Language Inference (NLI) task, i.e., judging whether a hypothesis is entailed by the premise. A lot of NLI datasets are available \cite{bowman-etal-2015-large, N18-1101, thorne-etal-2018-fever, nie-etal-2020-adversarial} and recent NLI models have achieved close-to-human-level performance. Hence, we use a pretrained NLI model and finetune it on some in-domain gold labels of SCUs' presence. Then, we replace humans with the finetuned model, so that the evaluation results are reproducible as long as the same model is used. Meanwhile, it can run automatically during development to guide model selection and the evaluation cost will be dramatically reduced. \citet{shapira2019crowdsourcing} propose \emph{LitePyramid} to simplify the standard Pyramid method via crowdsourcing.
Following but different from their work, 
we additionally automate the presence annotation, and hence we call our method \emph{Lite$^2$Pyramid}.   

\emph{Lite$^2$Pyramid} still requires human efforts to extract SCUs from reference summaries, and this step is usually considered to be more difficult. Early benchmarks, e.g., TAC \cite{DBLP:conf/tac/2008, DBLP:conf/tac/2009}, are small-sized with fewer than 100 examples in the evaluation set, for which it is already expensive to manually collect SCUs. However, current popular summarization datasets, e.g., CNN/DM \cite{hermann2015teaching}, contain more than 10K evaluation examples, and hence we want to simulate SCUs via an automatic method for such large-scale datasets. For this, we make use of Semantic Role Labeling (SRL) that can automatically decompose a sentence to semantic triplets, e.g., \emph{subject-verb-object}, and we take each triplet as a pseudo-SCU, which we call Semantic Triplet Unit (STU). Figure~\ref{fig:metric} illustrates the difference between SCUs and STUs. Although STUs do not always contain a single fact and some information might also be misrepresented, we find that it can reasonably simulate SCUs and lead to a fully automatic metric, \emph{Lite$^3$Pyramid}. 

Lastly, instead of using either all human-labeled SCUs or all automated STUs, we investigate balanced trade-offs in between, e.g., using half SCUs and half STUs. A naive way is to randomly sample some reference sentences and substitute their SCUs with STUs. However, we find it is unstable and sometimes even works worse than using all STUs. More reasonably, we design an \emph{active learning} \cite{settles2012active} inspired selection method to help decide which sub-parts of the dataset are more worthy of obtaining expensive SCUs for.
For this, we develop a regressor to predict the ``simulation easiness'' of each reference sentence: if a sentence is too complex to be well represented by STUs, we will ask humans to annotate SCUs for it; otherwise, we can apply automatic SRL. 
We call this method as \emph{Lite$^{2.x}$Pyramid}, since it provides a smooth, flexible transition from \emph{Lite$^2$Pyramid} to \emph{Lite$^3$Pyramid} and balances reliability with cost. 

To comprehensively evaluate the quality of metrics, we not only use 3 existing meta-evaluation datasets (TAC2008 \cite{DBLP:conf/tac/2008}, TAC2009 \cite{DBLP:conf/tac/2009}, REALSumm \cite{bhandari-etal-2020-evaluating}) but also newly collect \emph{PyrXSum} with 100 XSum \cite{narayan-etal-2018-dont} test examples plus summaries produced by 10 systems. Next, we compare our new metrics to 15 existing automatic metrics on these 4 meta-evaluation setups for both \emph{system-level} and \emph{summary-level} correlations with human Pyramid scores. We find that \emph{Lite$^2$Pyramid} consistently has the best summary-level correlations and is reliable as an out-of-the-box metric. \emph{Lite$^3$Pyramid} also mostly performs better or competitively. Lastly, the regressor-based \emph{Lite$^{2.x}$Pyramid} can help substantially reduce annotation efforts for only small correlation drops, e.g., on TAC2008, TAC2009, it trades off only 0.01 absolute summary-level Pearson correlation and 0 system-level correlation for 50\% SCU reduction.

\section{Related Works \& Background}
Each example in a summarization dataset contains single or multiple source document(s) and one or several human-written reference(s). 
System-generated summaries are evaluated by comparing them to the references (i.e., reference-based) or directly scored (i.e., reference-free). This evaluation process is critical and directly affects our development choices. 

\textbf{Human (or manual) evaluation} has been considered as the gold standard. Early benchmarks \cite{DBLP:conf/tac/2008, DBLP:conf/tac/2009} conducted three human evaluations: \emph{Responsiveness}, \emph{Linguistic Quality}, and \emph{Pyramid}. The first two ask humans to directly rate the overall responsiveness or linguistic quality on a Likert scale. Following this, some works collect ratings for different aspects, e.g., relevance, readability \cite{paulus2018deep, kryscinski2019neural, fabbri2021summeval}. However, these ratings may suffer from raters' subjectivity. Pyramid \cite{nenkova-passonneau-2004-evaluating} has been perceived as a more objective method, and it is reference-based. It has two steps: \emph{pyramid creation} and \emph{system evaluation}. In the first step, humans exhaustively find the Summary Content Unit (SCU) contributors from references, each contributor describes a single fact; contributors with the same meaning will be merged into one single SCU; then each SCU is weighted by how many contributors it has, equal to the number of references in which it is found. In the second step, each SCU has been manually checked its presence in the system summary; and the Pyramid score is the normalized sum of present SCUs' weights (essentially, a recall score). \citet{passonneau2010formal} normalize it by the total weight of the best possible summary. Recently, \citet{shapira2019crowdsourcing} propose \emph{LitePyramid}. It removes SCU merging and weighting, allowing SCUs of the same meaning to co-exist, and they show that the evaluation can be reliably conducted by crowdsourcing workers. 
\vspace{2pt}

\textbf{Automatic metrics} trade off the reliability of human evaluation for reproducibility, low cost, and fast speed. Many automatic metrics have been introduced, the majority of which are reference-based. Some metrics measure the n-gram overlap \cite{papineni2002bleu, lin2004rouge}, out of which ROUGE \cite{lin2004rouge} is the most widely adopted metric till today.
Some other works compute the similarity over n-gram graphs \cite{giannakopoulos2011autosummeng, giannakopoulos2008summarization} or distributions \cite{lin-etal-2006-information}.
Since exact n-gram matching is too rigid, METEOR \cite{banerjee2005meteor, denkowski2014meteor} provides flexibility by stemming, synonyms, etc., and recently, a few metrics enable ``soft'' matching through contextualized word embeddings \cite{zhao2019moverscore, clark2019sentence, zhang2019bertscore}. However, \citet{deutsch2020understanding} point out that the n-gram based metrics indicate more topic similarity than information overlap. 
Structural evaluation metrics have also been proposed beyond n-grams. BEwT-E \cite{tratz2008bewte} decomposes the system summary and the reference(s) into syntactic units and compute their similarities, and decomposed-ROUGE \cite{deutsch2020understanding} computes ROUGE for each syntactic category. APES \cite{eyal2019question} and QAEval \cite{deutsch2020towards} are QA-based metrics that assume similar answers will be obtained from similar system summaries and reference(s).

Automatic Pyramid methods have also been proposed \cite{yang2016peak, hirao-etal-2018-automatic, gao2019automated}. They usually decompose both the system summary and the references into smaller units (e.g., Elementary Discourse Units) and compare the two list of units. Differently, our \emph{Lite$^3$Pyramid} only decomposes the reference summaries to semantic triplet units (STUs), and we use NLI to judge the presence of each STU in the system summary, which is closer to the original Pyramid's procedure and leads to better correlations with human scores (refer to Section~\ref{sec:res}). \citet{peyrard2017learning} propose a learned metric, S3, that is trained to directly predict human Pyramid or Responsiveness scores based on ROUGE, FrameNet features, etc. \citet{sellam2020bleurt} propose a learned metric for machine translation, BLEURT, that finetunes a BERT \cite{devlin-etal-2019-bert} model with human ratings to directly predict the similarity score of a (reference, model translation) pair, and they show that it can also be successfully applied for WebNLG \cite{gardent2017webnlg} tasks. We are similar to both S3 and BLEURT in the way of learning to evaluate through finetuning NLP models with human labels.
\citet{xu2020fact} is distantly related to us in the way of representing texts by SRL, but it is used to weigh the content in the source document(s).
Besides, some reference-free metrics are introduced for summary quality estimation \cite{xenouleas2019sum,gao2020supert, vasilyev2020fill} or faithfulness evaluation \cite{durmus2020feqa, wang2020asking}. 

\textbf{Semi-automatic evaluation} is introduced by \citet{zhou2007semi}. They automatically decompose both system summary and reference(s) into semantic units and then ask humans to match/align the two lists of units. In contrast, our semi-automatic Lite$^2$Pyramid retains the reusable SCUs while automatically judges the SCUs' presence in the system summary (via NLI).

\section{Our Method}

\begin{figure*}
    \centering
    \includegraphics[width=1.0\textwidth]{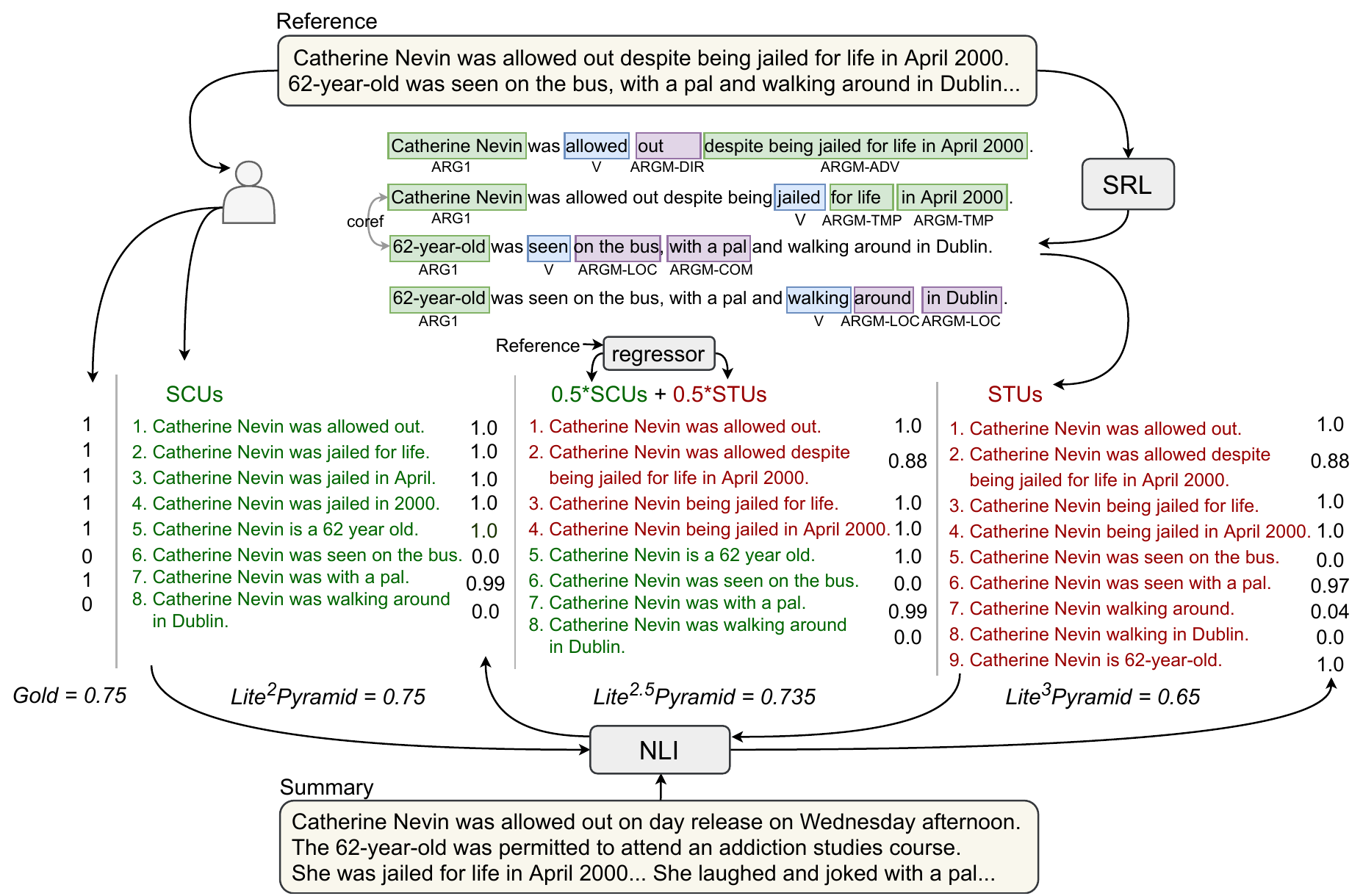}
    \vspace{-5pt}
    \caption{The illustration of our metrics. This data example is from REALSumm \cite{bhandari-etal-2020-evaluating} (we omit unnecessary content by `...'). For gold labels, `1' stands `present' and `0' stands `not present'. Other scores are the 2-class entailment probabilities, $p^{2c}(e)$, from our finetuned NLI model. 
    }
    \vspace{-5pt}
    \label{fig:metric}
\end{figure*}

\subsection{Lite$^2$Pyramid}
Lite$^2$Pyramid is a semi-automatic metric that retains human-labeled Summary Content Units (SCUs) to represent reference summaries of a data example $i$, i.e., $\{SCU_{ij}\}_{j=1}^{N_i}$, where $N_i$ is the total number of SCUs from all reference summaries. The original Pyramid \cite{nenkova-passonneau-2004-evaluating, passonneau2010formal} assumes there are multiple references available (e.g., TAC datasets \cite{DBLP:conf/tac/2008, DBLP:conf/tac/2009} have 4 references per example). Therefore, each SCU comes with weight, $\{w_{ij}\}_{j=1}^{N_i}$, representing the number of reference summaries in which the SCU is found. To evaluate a particular system summary $s_i$, the standard Pyramid method manually checks each SCU's presence, sums up the weights of present SCUs, and normalizes it:
\begin{equation}
    \label{eq:pyr}
 \text{Pyramid}_i =\frac{\sum_{j=1}^{N_i} w_{ij} \text{Presence}(SCU_{ij}, s_i)}{\text{the best possible score}}
 \end{equation}
The best possible score is the highest sum of weights the summary can obtain with the same number of present SCUs (details can be found in \cite{passonneau2010formal}). Differently, LitePyramid~\cite{shapira2019crowdsourcing} takes a union of SCUs from all reference summaries \emph{with duplication} (we use $SCU^*$ to distinguish it from the de-duplicated $SCU$ used above) 
and then samples the same number ($K$) of SCUs for every data example, hence:
$$\text{LitePyramid}_i =\frac{\sum_{j=1}^K \text{Presence}(SCU_{ij}^*, s_i)}{K}$$
Without weighting, this method also works in single-reference situations. Different from this method, we keep the exhaustive set (instead of a fixed-size sample) of SCUs for each example (also used by \citet{bhandari-etal-2020-evaluating}). Importantly, we replace human efforts of checking SCUs' presence with a Natural Language Inference (NLI) model $f_{\textsc{nli}}$'s 
entailment prediction. Using $e$ to denote entailment, our metric can be written as:
\begin{equation}
    \label{eq:lite2}
    \text{Lite}^2\text{Pyramid}_i =\frac{\sum_{j=1}^{N_i} w_{ij}f_{\textsc{nli}}(e|SCU_{ij}, s_i)}{\sum_{j=1}^{N_i} w_{ij}}
\end{equation}
Note that multiplying the weights and dividing by the sum of the weights is equal to repeating $SCU_i$ for $w_i$ times, which shows how we treat SCUs as an exhaustive set with duplication. For single-reference datasets (CNN/DM or XSum), the weights are all 1. Plus, the above equations all compute summary-level scores. To get one single score for the system, we simply take the average across examples, e,g., $\frac{1}{|D|}\sum_{i\in D}\text{Lite}^2\text{Pyramid}(s_i)$.

The $f_{\textsc{nli}}$ function can be implemented in four different ways, denoted as $p^{3c}$, $l^{3c}$, $p^{2c}$, $l^{2c}$, and explained below.
Following the standard 3-class setting of NLI tasks, the NLI model will predict whether the $SCU_{ij}$ is entailed by or neutral to or contradicted with the summary $s_i$. Hence, we can use either the output probability of entailment class $p^{3c}(e)$ or the predicted 1 or 0 entailment label $l^{3c}(e)$ as the function $f_{\textsc{nli}}$. However, existing NLI datasets \cite{bowman-etal-2015-large, williams2018broad, thorne-etal-2018-fever, nie-etal-2020-adversarial} have different data distributions and domains from the summarization data; hence models trained on these datasets may not perform well in judging the presence of SCUs. Therefore, we finetune the pretrained NLI model by human-labeled SCUs plus presence labels. Since humans only give 2-class labels (present or not present), we adapt the model to perform two-way classification. Specifically, we add up the logits of neutral ($n$) and contradiction ($c$) classes as the logit of the ``not present'' label: $p^{2c}(e) = \frac{exp(\text{logit}_e)}{exp(\text{logit}_e) + exp(\text{logit}_n + \text{logit}_c)}$. Again, we can use $p^{2c}(e)$ or $l^{2c}(e)$ as $f_{\textsc{nli}}$ after finetuning. In our experiments, we call the pretrained NLI model on NLI datasets as ``zero-shot'' because it has not seen summarization data. Empirically, we find that when using the zero-shot NLI model, $l^{3c}$ works best; while after finetuning, $p^{2c}$ usually works best.

\subsection{Lite$^3$Pyramid}
Lite$^3$Pyramid fully automates Lite$^2$Pyramid by simulating the human-annotated SCUs with automatic extracted semantic triplets. We use a Semantic Role Labeling (SRL) model \cite{carreras2005introduction, palmer2010semantic, he2017deep, shi2019simple} to achieve this goal. SRL determines the latent predicate-argument structure of a sentence, e.g., \emph{who did what to whom}. 
As shown in Figure~\ref{fig:metric}, the SRL model will identify several frames for each sentence, and each frame has one verb and a few arguments. For each frame, we keep the verb and any arguments before the verb unchanged, then we enumerate the arguments after
the verb to form a list of triplets as \{(ARG$_{before}$, V, ARG$_{after}^i$)\}$_{i=1}^M$, where $M$ is the number of arguments after the verb. We concatenate the three elements in each triplet to form a short sentence because a SCU is a short sentence and we want to resemble it as much as possible.
We call these short sentences Semantic Triplet Units (STUs).\footnote{Note that simple concatenation might not lead to grammatical sentences, but we expect the NLI model to be robust to small grammar errors. Additionally, we make a small fix in two cases: if the token before V in the original sentence is classified as a negation modifier, ARGM-NEG, or is a \emph{Be} verb, we add it to the STU sentence (e.g., for the 3rd STU in Figure~\ref{fig:metric}, we bring back ``being'' before ``jailed'').}   For example, as illustrated by Figure~\ref{fig:metric}, based on the 4 frames identified by SRL, we extract 9 STUs from the reference.

Since one entity can be referred to by pronouns or different names in the summary, we also apply Coreference Resolution \cite{lee-etal-2018-higher} 
to improve the simulation quality. As shown in Figure~\ref{fig:metric}, \emph{Catherine Nevin} and \emph{62-year-old} are identified as coreference, so we use \emph{Catherine Nevin} as the subjects of STUs and add an additional STU \emph{Catherine Nevin is 62-year-old}.\footnote{In practice, we use the name appeared first in the reference to unify the mentions in STUs and use the template ``name$_1$ is name$_n$'' to generate additional STUs.} In our experiments, we only apply coreference resolution for REALSumm
because empirically, on TAC datasets, we find applying it works worse than not applying; and PyrXSum has one-sentence summaries where coreference hardly appears.\footnote{Even for REALSumm, removing the coreference resolution step will only cause around 0.01 absolute correlation drops.} Although STUs seem to reasonably simulate SCUs for the example in Figure~\ref{fig:metric}, it has limitations, especially, when the sentence is syntactically complicated, e.g., with a lot of modifiers, clauses, complements (refer to Section~\ref{sec:res} for more discussions).

After we obtain the STUs from all reference summaries, we score a system summary $s_i$ by:
\begin{equation*}
\text{Lite}^3\text{Pyramid}_i =\frac{1}{{M_i}}\sum_{j=1}^{M_i} f_{\textsc{nli}}(e|STU_{ij}, s_i)
\end{equation*}
where $M_i$ is the total number of STUs. Note that there is no weight because we extract STUs from every reference summary and take a union, which allows STUs of the same meaning to co-exist. 

\subsection{Lite$^{2.x}$Pyramid}
As discussed so far, human-annotated SCUs are accurate yet expensive, 
whereas automatically extracted STUs are cheap yet sometimes erroneous. The next natural question is how to find a balance between them. One way is to randomly replace 50\% sentences' SCUs with STUs, but a more intuitive way is to make the decision based on the ``easiness'' of simulating the sentence's SCUs by STUs. If the sentence is unlikely to be well represented by STUs, we can ask humans to label SCUs for it; otherwise, we can use STUs to reduce cost. This is similar to how \emph{active learning} \cite{settles2012active} chooses which training examples to collect human labels for. We define simulation easiness as the average simulation accuracy of each SCU. ROUGE-1-F1 (R1$_{\text{F1}}$) \cite{lin2004rouge} is used to measure the simulation accuracy: $\text{Acc}_j = max_m \text{R1}_{\text{F1}}(SCU_j, STU_m)$. Then, the easiness of a sentence with $N_{sent}$ SCUs is written by $\text{Easiness}_{sent} = \frac{1}{N_{\text{sent}}}\sum_{j=1}^{N_{sent}} \text{Acc}_j$. The higher the easiness score is, the more accurately the STUs resemble SCUs.

After we obtain these gold easiness scores, we want to train a regressor to predict the score based on sentence complexity features. As we mentioned above, the sentence's syntax can indicate its simulation difficulty. Therefore, we get the Constituency Parsing tree \cite{joshi-etal-2018-extending} of each sentence and define the following features: (1) sentence length; (2) linearized parsing tree length; (3) parsing tree depth; (4) sentence length / parsing tree depth; (5) the counts for each of the 65 non-terminal tokens (e.g., NNP). In total, we represent each sentence with a 69-dim feature vector. Then, we train an XGBoost \cite{chen2016xgboost} regressor to predict the simulation easiness by minimizing the mean squared errors. Given this regressor, we propose to replace top $0.x$ scored sentences' SCUs  with STUs, leading to Lite$^{2.x}$Pyramid. For example, Lite$^{2.5}$Pyramid (illustrated in Figure~\ref{fig:metric}) means that we use STUs for the top 50\% scored sentences and use SCUs for the other half.

\section{Evaluation}
\label{sec:eval}
\paragraph{Correlation with human scores.}

Following the standard meta-evaluation strategies used in previous works \cite{peyrard2017learning, bhandari-etal-2020-evaluating, deutsch2020towards}, we evaluate metrics by two types of correlation with gold human scores. 
\vspace{2pt}

\noindent\textbf{System-level} correlation aims to evaluate \emph{how well the metric can compare different summarization systems?} 
We denote the correlation measure as $K$, human scores as $h$, the metric as $m$, and generated summaries as $s$. We assume there are $N$ examples and $S$ systems in the mete-evaluation dataset. Then, the system-level correlation is defined as:
\begin{align*}
    K^{sys}_{m, h} = K(& [\frac{1}{N}\sum_{i=1}^N m(s_{i1}), ..., \frac{1}{N}\sum_{i=1}^N m(s_{iS})], \\
    & [\frac{1}{N}\sum_{i=1}^N h(s_{i1}), ..., \frac{1}{N}\sum_{i=1}^N h(s_{iS})])
\end{align*}
\noindent\textbf{Summary-level} correlation answers \emph{if the metric can reliably compare summaries generated by different systems for the same document(s)}. Using the same notations, this correlation is written by:
\begin{align*}
    K^{sum}_{m, h} = \frac{1}{N}\sum_{i=1}^N K(& [m(s_{i1}), ..., m(s_{iS})], \\ & [h(s_{i1}), ..., h(s_{iS})])
\end{align*}
We use Pearson $r$ or Spearman $\rho$ as the correlation measure $K$. Pearson measures linear correlation while Spearman measures ranking correlation.

\paragraph{Metrics for comparison.}
Taking advantage of SacreROUGE \cite{deutsch-roth-2020-sacrerouge}, we compare our metrics to 15 existing metrics: \emph{ROUGE-1}, \emph{ROUGE-2}, and \emph{ROUGE-L} \cite{lin2004rouge}, \emph{AutoSummENG} \cite{giannakopoulos2008summarization}, \emph{METEOR} \cite{banerjee2005meteor, denkowski2014meteor}, \emph{BEwT-E} \cite{tratz2008bewte}, \emph{S3 (pyr)} and \emph{S3 (resp)} \cite{peyrard2017learning}, \emph{PyrEval}'s \emph{qual}ity and \emph{comp}rehensive scores \cite{gao2019automated},\footnote{PyrEval is the latest automatic Pyramid metric and is shown to be better than \citet{yang2016peak}.} \emph{BERTScore} and \emph{BERTScore (idf)} \cite{zhang2019bertscore}, \emph{MoverScore} \cite{zhao2019moverscore}, \emph{QAEval (EM)} and \emph{QAEval (F1)} \cite{deutsch2020towards}. For metrics that have \emph{precision/recall/F1}, we report \emph{recall} because Pyramid is essentially recall-based. Note that PyrEval only supports multi-reference situations.\footnote{\url{https://github.com/serenayj/PyrEval/issues/11}} See the complete descriptions of these metrics in Appendix~\ref{appendix:metrics}.

\paragraph{Data.} We evaluate human-metric correlations on three existing English meta-evaluation datasets: \emph{TAC2008} \cite{DBLP:conf/tac/2008}, \emph{TAC2009} \cite{DBLP:conf/tac/2009}, \emph{REALSumm} \cite{bhandari-etal-2020-evaluating}. TAC08 contains 96/58 examples/systems and TAC09 has 88/55 examples/systems. We compute the correlations with their official Pyramid scores (Equation~\ref{eq:pyr}).\footnote{We find that the exhaustive set based computation (replacing $f_{\textsc{nli}}$ in Equation~\ref{eq:lite2} by gold labels) has close to perfect correlation with TAC's official scores. REALSumm also use this computation as reflected by the gold score in Figure~\ref{fig:metric}.} REALSumm has 100 CNN/DM \cite{hermann2015teaching} test examples and 25 systems. They label SCUs by themselves and collect SCU-presence labels on Amazon Mechanical Turk (AMT). Both TAC and CNN/DM have long and extractive summaries. To complete our evaluation, we newly collect an English meta-evaluation dataset \emph{PyrXSum} for 100 XSum \cite{narayan-etal-2018-dont} (has short and abstractive summaries) testing examples. Following REALSumm, we (authors) manually label SCUs and collect SCU-presence labels for summaries generated by 10 systems\footnote{Fast Abs RL \cite{chen2018fast}, PtGen \cite{see2017get}, ConvS2S and T-ConvS2S \cite{narayan-etal-2018-dont}, TransAbs and BertAbs and BertExtAbs \cite{liu2019text}, T5 \cite{raffel2020exploring}, BART \cite{lewis2020bart}, PEGASUS \cite{zhang2020pegasus}} on AMT.
We collect 4 responses per summary (100 * 10 * 4 HITs) and filter responses from a noisy worker. We use the majority vote to label each SCU's presence and break ties by “not present”. See more data collection details of PyrXum in Appendix~\ref{appendix:pyrxsum}.

\paragraph{Models.} We use the pretrained RoBERTa-large \cite{liu2019roberta} based NLI model released by \citet{nie-etal-2020-adversarial}, which has been trained on multiple NLI datasets.
We continually finetune this model with the gold SCUs plus SCU-presence labels always for 2 epochs. For SRL, Coreference Resolution, and Constituency Tree Parser, we use the out-of-the-box tools provided by AllenNLP \cite{gardner2018allennlp, shi2019simple, lee-etal-2018-higher, joshi-etal-2018-extending}. See the complete implementation details in Appendix~\ref{appendix:expdetails}.

\section{Results}
\label{sec:res}
\begin{table*}[t]
\centering
\small
\resizebox{0.99\textwidth}{!}{%
\begin{tabular*}{1.03\textwidth}{l cc cc cc cc c cc cc cc cc}
\toprule
& \multicolumn{8}{c}{System-level} &  & \multicolumn{8}{c}{Summary-level} \\
\cmidrule(lr){2-9} \cmidrule(lr){11-18}
& \multicolumn{2}{c}{TAC08} & \multicolumn{2}{c}{TAC09} & \multicolumn{2}{c}{RealSumm} & \multicolumn{2}{c}{PyrXSum} &  & \multicolumn{2}{c}{TAC08} & \multicolumn{2}{c}{ TAC09} & \multicolumn{2}{c}{RealSumm} & \multicolumn{2}{c}{PyrXSum} \\
Metrics & $r$ & $\rho$ & $r$ & $\rho$ & $r$ & $\rho$ & $r$ & $\rho$ & & $r$ & $\rho$ & $r$ & $\rho$ & $r$ & $\rho$ & $r$ & $\rho$ \\
\midrule
ROUGE-1 & .87 & .87 & .91 & .86 & .82 & .83 & .92 &.90 & & .62 & .61 & .69  & .63 & .53 & .50 & .52 & .50\\
ROUGE-2 & .90 & .90 & .92 & .90 & .84 & .82 & .93& .91& & .63 & .62 & .71 & .64 & .46 & .43 & .53 & .51 \\
ROUGE-L & .87 & .86 & .93 & .87 & .83 & .81 & \underline{.94} & \underline{\bf.92} & & .57 & .55 & .66 & .59 & .46 & .42 & .52 & .51 \\
AutoSummENG & .90 & .89 & .91 & .89 & .53 & .51 & .92 & .91  & & .65 & .64 & .71 & .64 & .34 & .34 & .56 & .53 \\
METEOR & .90 & .89 & .93 & .88 & .84 & .84 & \underline{.94} & .89 & & .65 & .64 & .73 & .68 & .54 & .49 & \underline{.58} & \underline{.56}\\
BEwT-E & .92 & \underline{.91} & .95 & .92 & .83 & .84 & .93 & .86 & & .66 & .65 & .75 & .68 & .47 & .45 & .54 & .52 \\
S3 pyr & .90 & .89 & .95 & .89 & .86 & .85 & \underline{.94} & .89 & & .66 & .65 & .75 & .68 & .54 & .50 & .57 & .54\\
S3 resp & .91 & .91 & .94 & .90 & .86 & .86 & \underline{.94} & .90 & & .67 & .65 & .74 & .68 & .52 & .48 & .57 & .54 \\
PyrEval qual & .83 & .81 & .88 & .80 & - & - & - & - & & .40 & .39 & .49 & .44 & - & - & - & - \\
PyrEval comp & .83 & .80 & .90 & .79 & - & - & - & - & & .41 & .40 & .53 & .45 & - & - & - & -\\
BertScore & .88 & .87 & .90 & .90 & .73 & .77 & .92 & .89 & & .61 & .60 & .70 & .65 & .48 & .46 & .57 & .54 \\
BertScore (idf) & .89 & .88 & .91 & .90 & .73 & .78 & .93 & .90 & & .62 & .61 & .71 & .66 & .48 & .46 & \underline{.58} & .55\\
MoverScore & .91 & .89 & .95 & .90 & .40 & .31 & .92 & .91 & & .64 & .63 & .73 & .68 & .39 & .36 & .57 & .54\\
QAEval EM & .83 & .81 & .85 & .83 & .61 & .51 & .86 & .85 & & .48 & .48 & .64 & .55 & .28 & .27 & .29 & .27 \\
QAEval F1 & .89 & .87 & .90 & .87 & .72 & .65 & .90 & .83 & & .61 & .60 & .70 & .63 & .38 & .35 & .46 & .42 \\
\midrule
Lite$^3$Pyramid & \underline{.93} & \underline{.91} & \underline{\bf.97} & \underline{.93} & \underline{.89} & \underline{.87} & .89 & .86 & & \underline{.71} & \underline{.69} & \underline{.78} & \underline{.73} & \underline{.57} & \underline{.53} & .51 & .48 \vspace{4pt}\\
Lite$^{2.5}$Pyramid & \bf .95& \bf .93 & \bf.97 & \bf.94 & \bf.90 & \bf .88 & .92 & .87& & .76& .75&.82 & .77 & .62 & .57 & .64 & .59 \\
Lite$^2$Pyramid & \bf .95 & \bf .93 & \bf.97 & \bf.94 & .89 & .86 & .95 & \bf.92 & & \bf.77 & \bf .76 & \bf.83 & \bf.78 & \bf.64 & \bf.60 & \bf .74 & .66\\
Lite$^2$Pyramid-0 & .86 & .83 & .95 & .88 & .86 & .82 & \bf.96& \bf.92& & .62 & .61 & .74 & .68 & .56 & .53 & .73 & \bf .72 \\
\bottomrule
 \end{tabular*}
 }
 \vspace{-5pt}
\caption{5-fold (split by examples) cross-validation results. In each column, the \textbf{bold} numbers are the best and the \underline{underline} numbers are the best out of automatic metrics. All Lite$^2$Pyramid-0 numbers are based on $f_{\textsc{nli}}=l^{3c}$, while all other numbers of our metrics are based on $f_{\textsc{nli}}=p^{2c}$.} 
\label{table:result_example}
\vspace{-8pt}
\end{table*}

\subsection{Human-Metric Correlation Results}
Since we find that finetuning the NLI model with in-domain presence labels is greatly beneficial, following \citet{peyrard2017learning}, we evaluate by 5-fold cross-validation. For each dataset, we split it into 5 folds, finetune the NLI model and train the regressor on 4 folds, test on the left one, and repeat for 5 times. We report the 5-fold average correlations of both our metrics and the 15 metrics we compare to for fair comparison. Instead of random splitting, we split the data \emph{by examples} or \emph{by systems}, aiming to check the generalizability across examples or systems. E.g., if we split REALSumm by examples, each fold has summaries of 20 examples; when split by systems, each fold has summaries generated by 5 systems.

Table~\ref{table:result_example} shows our 5-fold (split \emph{by examples}) cross-validation results. Firstly, it can be observed that our \textbf{Lite$^2$Pyramid} always has the best or close to the best correlations; especially, it has 0.08 to 0.16 higher summary-level correlations than the best metrics we compare to. It demonstrates the advantage of semi-automatic evaluation which dramatically improves reliability without losing reproducibility. Meanwhile, it indicates that the finetuned NLI model can generalize to new data examples and works reasonably well as a proxy of human judgment. In contrast, Lite$^2$Pyramid-0, which uses a non-finetuned NLI model, usually works greatly worse than Lite$^2$Pyramid, which indicates the importance of in-domain finetuning. It is surprising that Lite$^2$Pyramid-0 works better than or similar to Lite$^2$Pyramid on PyrXSum. We conjecture that because our PyrXSum is relatively small-size, the finetuning will not make big difference. 

Secondly, our \textbf{Lite$^3$Pyramid} has the best correlations comparing to the other automatic metrics, except for PyrXSum; again, its advantage is more prominent on summary-level correlation (around 0.03 to 0.05 better). Its failure in PyrXSum is caused by the limitation of SRL. XSum's reference summary sentences usually have a lot of modifiers, adverbial phrases/clauses, or complements, which increases the difficulty of decomposing it into STUs. E.g., for the summary \emph{``Netherlands midfielder Wesley Sneijder has joined French Ligue 1 side Nice on a free transfer''}, human annotates the following 5 SCUs: \emph{``Wesley Sneijder is a midfielder''}, 
\emph{``Wesley Sneijder comes from Netherlands''}, \emph{``Wesley Sneijder has joined French Ligue 1 side''}, \emph{``Wesley Sneijder has joined Nice''}, and \emph{``Wesley Sneijder has been on a free transfer''}. However, since SRL frames are centered around verbs, it can only extract two STUs: \emph{``Netherlands midfielder Wesley Sneijder joined French Ligue 1 side Nice''} and \emph{``Netherlands midfielder Wesley Sneijder joined on a free transfer''}. On average, human labels 4.8 SCUs per PyrXSum summary, however, the number is only 2.8 for STUs. Hence, a better semantic unit decomposer needs to be designed to improve Lite$^3$Pyramid's accuracy. 

\begin{figure}
    \centering
    \includegraphics[width=0.44\textwidth]{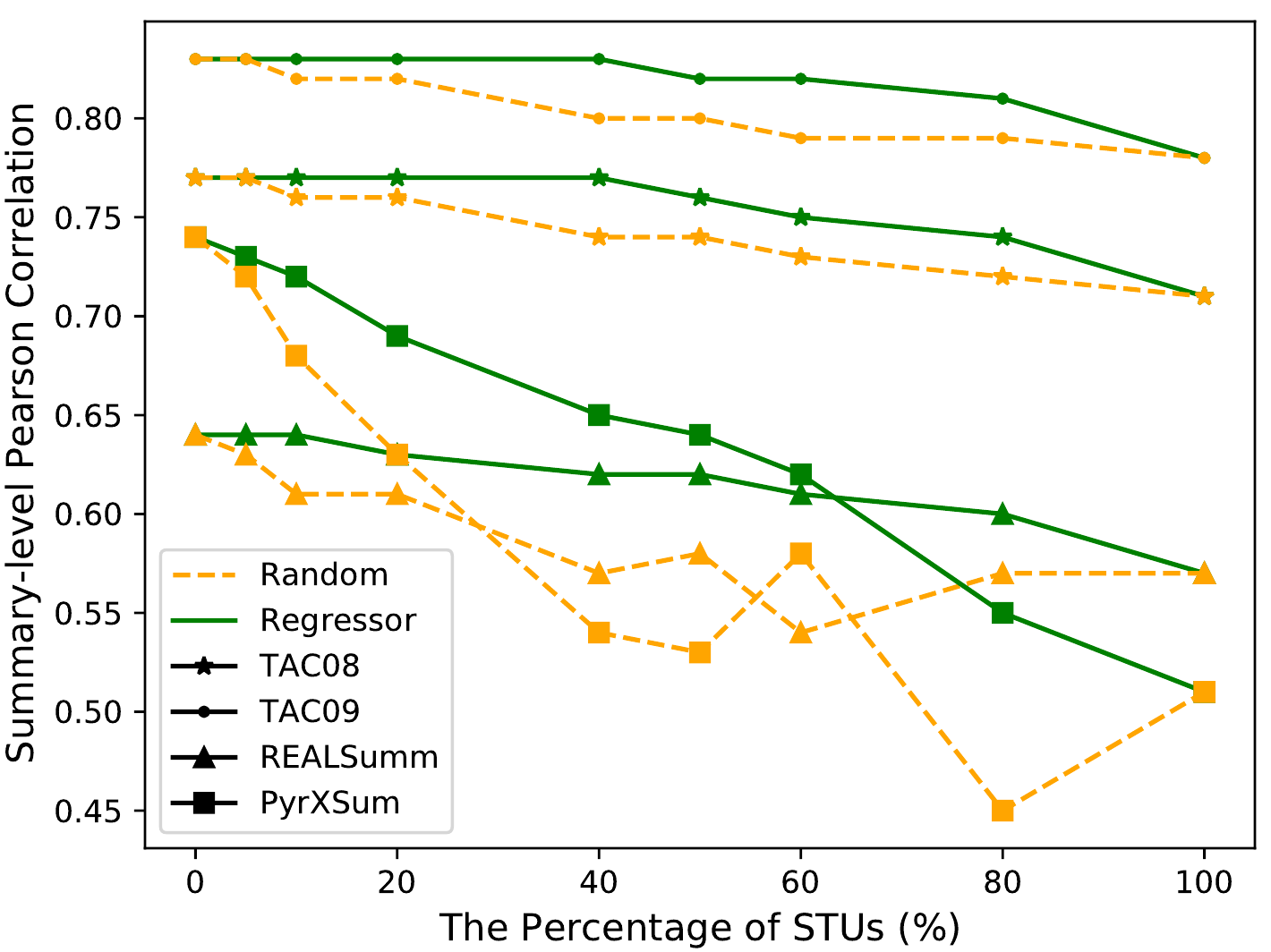}
    \vspace{-5pt}
    \caption{Lite$^{2.x}$Pyramid curves and its comparison to replacing \emph{random} sentences' SCUs with STUs.}
    \vspace{-10pt}
    \label{fig:lite2x}
\end{figure}

Lastly, \textbf{Lite$^{2.x}$Pyramid} alleviates the problem mentioned above by deferring complex sentences to humans to annotate SCUs for. As shown in Table~\ref{table:result_example}, Lite$^{2.5}$Pyramid, which saves half human effort by substituting 50\% sentences' SCUs with STUs, always has correlation reduction less than half of the difference between Lite$^2$Pyramid and Lite$^3$Pyramid and sometimes even has better system-level correlations than Lite$^2$Pyramid. The full Lite$^{2.x}$Pyramid curves are shown in Figure~\ref{fig:lite2x}, where the x-axis is the percentage of STUs (the higher means the fewer human efforts involved) and the y-axis is the summary-level Pearson correlation (Figure~\ref{fig:lite2x_sys} in Appendix shows system-level correlations). We can see that our Lite$^{2.x}$Pyramid offers a smoothing transition from semi-automatic Lite$^{2}$Pyramid to automatic Lite$^{3}$Pyramid. More importantly, compared to randomly selecting sentences (yellow dash lines), our regressor-based selection achieves a slower correlation reduction, i.e., saving the same amount of human effort our method can retain higher metric quality. Plus, this curve gives people flexible choices per their budget.

Due to space limitations, the 5-fold (split \emph{by systems}) cross-validation results are in Table~\ref{table:result_system} of Appendix. The same trends are mostly observed. Lite$^2$Pyramid still has 0.06 to 0.21 higher summary-level correlations across all datasets. Lite$^3$Pyramid achieves the best or competitive correlations comparing to other automatic metrics except for the system-level correlations on REALSumm and PyrXSum. And, Lite$^{2.x}$Pyramid also nicely bridges Lite$^2$Pyramid and Lite$^3$Pyramid and works better than random replacement. However, differently, Lite$^2$Pyramid does not get the best system-level correlations on REALSumm and PyrXSum, which may indicate the bigger generalization challenge across different systems. 

\noindent\textbf{Takeaway:} Lite$^2$Pyramid consistently has the best summary-level correlations and the best system-level correlations in most cases. The automatic  Lite$^3$Pyramid also mostly works better than other automatic metrics. Lite$^{2.x}$Pyramid provides flexible and balanced degrees of automation per budget.

\subsection{Out-of-the-Box Generalization}

\begin{table*}[t]
\centering
\small
\resizebox{0.95\textwidth}{!}{%
\begin{tabular*}{0.97\textwidth}{l l cc cc cc cc cc cc}
\toprule
& & \multicolumn{6}{c}{System-level} & \multicolumn{6}{c}{Summary-level} \\
\cmidrule(lr){3-8} \cmidrule(lr){9-14}
& &\multicolumn{2}{c}{ TAC09} & \multicolumn{2}{c}{REALSumm} & \multicolumn{2}{c}{PyrXSum} & \multicolumn{2}{c}{TAC09} & \multicolumn{2}{c}{REALSumm} & \multicolumn{2}{c}{PyrXSum} \\
& Metrics & $r$ & $\rho$ & $r$ & $\rho$ & $r$ & $\rho$ & $r$ & $\rho$ & $r$ & $\rho$ & $r$ & $\rho$ \\
\midrule
& ROUGE-1 & .93 & .89 & .91 & .92 & .98 & .96 & .69 & .63 & .53 & \underline{.50} & .52 & .50 \\
& ROUGE-2 & .94 & .95 & .96 & \underline{\textbf{.95}} & \underline{\textbf{.99}} & .95 & .71 & .64 & .46 & .43 & .53 & .51 \\
& ROUGE-L & .96 & .92 & .94 & \underline{\textbf{.95}} & \underline{\textbf{.99}} & .95 & .66 & .59 & .46 & .42 & .52 & .51 \\
& AutoSummENG & .93 & .93 & .59 & .60  & .97 & .94 & .71 & .64 & .34 & .34 & .56 & .53 \\
& METEOR & .95 & .91 & .94 & .92 & \underline{\textbf{.99}} & .98 & .73 & .68 & \underline{.54} & .49 & \underline{.58} & \underline{.56} \\
& BEwT-E & .97 & .96 & .91 & .89 & \underline{\textbf{.99}} & .98 & .75 & .68 & .47 & .45 & .54 & .52 \\
& S3 pyr & .97 & .92 & .96 & .94 & \underline{\textbf{.99}} & \underline{\textbf{.99}} & .75 & .67 & \underline{.54} & \underline{.50} & .57 & .54 \\
& S3 resp & .96 & .94 & \underline{\textbf{.97}} & \underline{\textbf{.95}} & \underline{\textbf{.99}} & .98 & .74 & .68 & .52 & .48 & .57 & .54 \\
& PyrEval qual & .94 & .90 & - & - & - & - & .49 & .44 & - & - & - & -\\
& PyrEval comp & .95 & .86 & - & - & - & - & .53 & .45 & - & - & - & -\\
& BertScore & .92 & .94 & .79 & .83 & .97 & .90 & .70 & .65 & .48 & .46 & .57 & .54 \\
& BertScore (idf) & .93 & .95 & .79 & .83 & .97 & .90 & .71 & .66 & .48 & .46 & .58 & .55 \\
& MoverScore & .97 & .92 & .44 & .32 & .98 & .84 & .74 & .68 & .39 & .36 & .57 & .54 \\
& QAEval EM & .88 & .94 & .88 & .86 & .95 & .95 & .64 & .55 & .28 & .27 & .29 & .27 \\
& QAEval F1 & .93 & .95 & .91 & .89 & .95 & .84 & .70 & .63 & .38 & .35 & .46 & .43\\
\midrule
TAC08& Lite$^3$Pyramid & \underline{\textbf{.99}} & \underline{.97} & .92 & .93 & .97 & .90 & \underline{.78} & \underline{.72} & .53 & .48 & .56 & .53 \\
 & Lite$^{2.5}$Pyramid & \textbf{.99} & .97 & .92 & .92 & .98 & .95 & .82 & .77 & .58 & .54 & .66 & .61 \\
 & Lite$^2$Pyramid & \textbf{.99} & \textbf{.98} & .94 & \textbf{.95} & \textbf{.99} & \textbf{.99} & \textbf{.83} & \textbf{.78} & \textbf{.61} & \textbf{.57} & \textbf{.71} & \textbf{.66}\\
\midrule
 TAC08 & Lite$^3$Pyramid &- & -& .94 & \underline{\textbf{.95}} & .97 & .88 & -&- & .52 & .49 & .56 & .53\\
 +TAC09 & Lite$^{2.5}$Pyramid & - & - & .93 & \textbf{.95} & .97 & .96 & -& -& .57 & .53 & .66 & .60 \\
 & Lite$^2$Pyramid & - & - & .94 & \textbf{.95} & \textbf{.99} & .98 &- &- & .59 & .56 & \textbf{.71} & .65 \\
\midrule
TAC08 & Lite$^3$Pyramid & -& -&- &- & .97 & .88 & -& -& -& -& .50 & .44\\
+TAC09 & Lite$^{2.5}$Pyramid & -&- & -& -& .98 & .94 & -& -& -& -& .60 & .55\\
+REALSumm& Lite$^2$Pyramid & -& -& -&- & \textbf{.99} & .94 & -& -&- &- & .70 & .64\\
\bottomrule
 \end{tabular*}
 }
 \vspace{-5pt}
\caption{Out-of-the-box generalization results. In each column, the \textbf{bold} numbers are the best and the \underline{underline} numbers are the best out of automatic metrics. }
\label{table:out_of_box}
\vspace{-8pt}
\end{table*}

We release the finetuned NLI models and the pretrained sentence regressors for future usage, so that they will work as out-of-the-box evaluation metrics for any summarization tasks. Then, a natural question to ask is \emph{how will the metrics perform on a new summarization task?} To better estimate the out-of-the-box performance, we simulate out-of-the-box situations by training the NLI model and the regressor on some dataset(s) and then evaluate metrics on the other dataset(s). For example, in the last big row (starting with TAC08+TAC09+REALSumm) of Table~\ref{table:out_of_box}, we finetune the NLI model and train the regressor on the entire TAC08+TAC09+REALSumm data then evaluate our metrics on PyrXSum only. Meanwhile, we also compare to other metrics. Different from the numbers in Table~\ref{table:result_example}, numbers in Table~\ref{table:out_of_box} are calculated on the entire meta-evaluation set instead of the average of 5 folds.

It can be observed from Table~\ref{table:out_of_box} that our Lite$^2$Pyramid retains its advantage in most out-of-the-box situations, especially for summary-level correlation. Though Lite$^3$Pyramid does not always outperform the best metrics, it stays competitive. In addition, Lite$^{2.5}$Pyramid retains its feature of trading off less than 50\% correlation for saving 50\% human effort. 
Surprisingly, learning from more data does not perform better: for PyrXSum, learning from all three other datasets (TAC08+TAC09+REALSumm) gets significantly worse performance than learning from TAC08 only or TAC08+TAC09. We conjecture that the difference between REALSumm (originated from CNN/DM \cite{hermann2015teaching}) and PyrXSum (originated from XSum \cite{narayan-etal-2018-dont}) leads to a ``distribution shift'', which causes the performance drop. Besides, though new metrics have been proposed, ROUGE is still the dominant evaluation metric in the summarization literature. However, based on our comparison, ROUGE is not the best evaluation choice in most cases, while METEOR \cite{banerjee2005meteor} and the learning-based metric, S3 \cite{peyrard2017learning}, have fairly good correlations with human judgment. Overall, our automatic Lite$^3$Pyramid is on a par with them, having the best performance in 4 cases (4 underline scores in Table~\ref{table:out_of_box}).

\noindent\textbf{Takeaway:} When evaluating for a new summarization task with human-labeled SCUs, one could expect that Lite$^2$Pyramid is reliably trustworthy and should be the top choice. Lite$^3$Pyramid is also a fairly good choice for fully automatic evaluation. Finally, our pretrained regressor can guide people on which data examples are more worthy of spending manual effort on annotating SCUs. 

\noindent\textbf{Speed:} Since SCUs' collection or STUs' extraction can be treated as data processing steps, the main speed bottleneck is running the NLI model. When a single TITAN V GPU is available, it takes around 2.5 minutes to evaluate 500 REALSumm (i.e., CNN/DM) examples. 

\noindent\textbf{Usage:} We provide the support of our metrics through our github repository and we will also incorporate it within the SacreROUGE library \cite{deutsch-roth-2020-sacrerouge}.

\section{Conclusion}
We propose to combine manual effort and automation for summary evaluation. We introduce a semi-automatic Lite$^2$Pyramid that gains reproducibility by replacing part of human effort with an NLI model. Following it, an automatic Lite$^3$Pyramid is proposed through decomposing references by SRL. Plus, we propose a simple yet effective regressor to decide which sentences are more worthy of labeling SCUs for, leading to flexible transition metrics, Lite$^{2.x}$Pyramid. Evaluating on four meta-evaluation datasets and comparing to 15 other automatic metrics, Lite$^2$Pyramid consistently has the best summary-level correlations; Lite$^3$Pyramid also performs better or competitively; and Lite$^{2.x}$Pyramid offers flexible degrees of automation, and its regressor will provide useful or expense-saving guidance for future datasets.

\section*{Acknowledgments}
We thank the reviewers for their helpful comments. We thank Xiang Zhou for useful discussions and thank Steven Chen for proofreading SCUs for PyrXSum. 
This work was supported by NSF-CAREER Award 1846185. 

\bibliography{custom}
\bibliographystyle{acl_natbib}

\appendix
\section{Appendix}
\label{sec:appendix}

\subsection{Metrics for Comparison}
\label{appendix:metrics}

\noindent\textbf{ROUGE-1}, \textbf{ROUGE-2}, and \textbf{ROUGE-L} \cite{lin2004rouge} are based on n-gram overlap and are widely used in summarization literature till today.

\noindent\textbf{AutoSummENG} \cite{giannakopoulos2008summarization} uses n-gram graphs to compare the system summary to the reference(s).

\noindent\textbf{METEOR} \cite{banerjee2005meteor, denkowski2014meteor} computes similarity through text alignment and uses stem, synonyms, paraphrases to allow more flexible matching. 

\noindent\textbf{BEwT-E} \cite{tratz2008bewte} decomposes summary into syntactic units and computes the similarity based on those units. 

\noindent\textbf{S3} \cite{peyrard2017learning} is a learned metric trained on TAC2008/2009 datasets to predict human Pyramid  (\textbf{pyr}) or Responsiveness (\textbf{resp}) scores. 

\noindent\textbf{PyrEval} \cite{gao2019automated} automate Pyramid by simulating SCUs through Emergent Discovery of Units of Attraction. It returns four scores. Empirically, we find that \textbf{qual}ity and \textbf{comp}rehensive work better, so we only keep these two in our result tables. Note that it only supports multi-reference situations because it retains SCUs' weighting step.

\noindent\textbf{BERTScore} \cite{zhang2019bertscore} aligns unigrams between two texts through the contextualized word embeddings from BERT \cite{devlin-etal-2019-bert}. We also compare to \textbf{BERTScore (idf)} that down-weights unigrams with high document frequency. 

\noindent\textbf{MoverScore} \cite{zhao2019moverscore} also uses contextualized word embeddings. Differently, they minimize the ``transportation cost'' between two texts.

\noindent\textbf{QAEval} \cite{deutsch2020towards} leverages Question Answering to evaluate the similarity of two texts, i.e., if they have the same meaning, the same answer should be inferred from them for the same question. They use either Exact Match (\textbf{EM}) or F1 (\textbf{F1}) to evaluate answer similarity.

\subsection{PyrXSum}
\label{appendix:pyrxsum}
Both TAC08/09 \cite{DBLP:conf/tac/2008, DBLP:conf/tac/2009} and REALSumm \cite{bhandari-etal-2020-evaluating} (examples from CNN/DM \cite{hermann2015teaching}) have long and extractive summaries. As a complementary, we collect a new meta-evaluation dataset, PyrXSum, for XSum \cite{narayan-etal-2018-dont} which contains short and abstractive summaries. We random sample 100 examples from XSum's testing set. Then, following \citet{bhandari-etal-2020-evaluating}, we (authors) annotate Semantic Content Units (SCUs) for reference summaries of the 100 examples. After annotation, another non-author native English speaker is invited to double-check the annotated SCUs and give improvement suggestions. Finally, we annotate 2 to 11 SCUs per reference; on average, there are 4.8 SCUs per reference. 

\begin{table*}[t]
\centering
\small
\resizebox{0.97\textwidth}{!}{%
\begin{tabular*}{1.08\textwidth}{lcccccccccc}
\toprule
Model & Fast Abs RL & PtGen & ConvS2S & T-ConvS2S & TransAbs & BertAbs & BertExtAbs & T5 & BART &  PEGASUS \\
\midrule
R2 & 7.02 & 9.68 & 11.58 & 11.46 & 10.85 & 15.63 & 17.68 & 21.01 & 23.96 & 26.23 \\
Pyramid & 0.09 & 0.09 & 0.12 & 0.12 & 0.07 & 0.19 & 0.22 & 0.29 & 0.31 & 0.31 \\
\bottomrule
 \end{tabular*}
 }
 \vspace{-5pt}
\caption{The ROUGE-2 (R2) and gold Pyramid scores obtained by 10 systems on the 100 XSum testing examples.}
\label{table:pyrxsum}
\vspace{-10pt}
\end{table*}

\begin{figure*}
    \centering
    \includegraphics[width=0.97\textwidth]{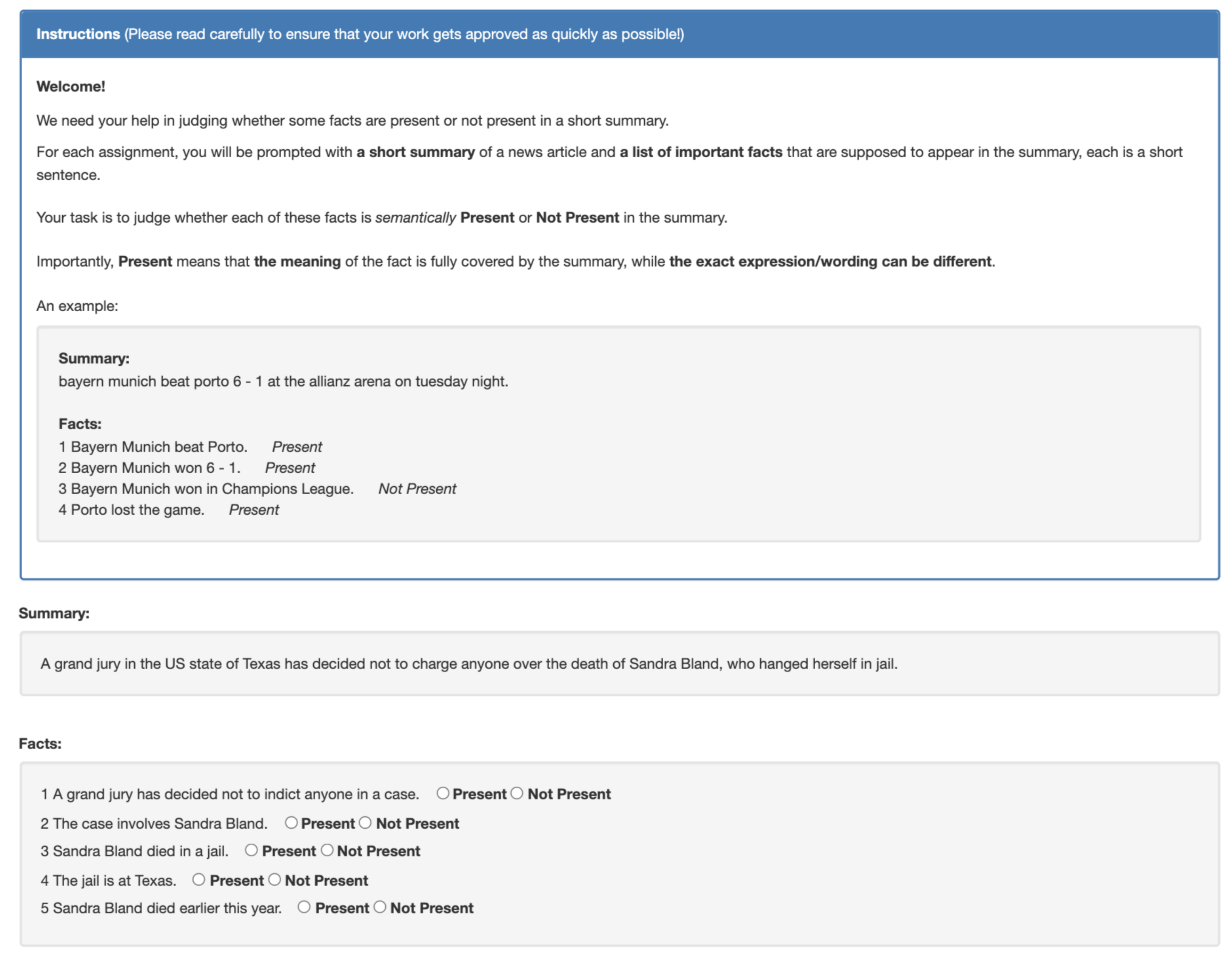}
    \vspace{-5pt}
    \caption{The Amazon Mechanical Turk user interface for collecting human labels of SCUs' presence.}
    \vspace{-12pt}
    \label{fig:pyrxsum}
\end{figure*}
Next, we obtain model generated summaries for these 100 examples from 10 abstractive summarization systems: Fast Abs RL \cite{chen2018fast}, PtGen \cite{see2017get}, ConvS2S and T-ConvS2S \cite{narayan-etal-2018-dont}, TransAbs and BertAbs and BertExtAbs \cite{liu2019text}, T5 \cite{raffel2020exploring}, BART \cite{lewis2020bart}, and PEGASUS \cite{zhang2020pegasus}. We do not include extractive summarization systems because XSum is known to be extremely abstractive and even oracle extractive method has low performance \cite{narayan-etal-2018-dont}. For Fast Abs RL, we use their open-source code\footnote{\url{https://github.com/ChenRocks/fast_abs_rl}} to train a model on XSum training set and get its generations for these 100 examples. We directly use the model outputs of PtGen, ConvS2S, and T-ConvS2S, released by \citet{narayan-etal-2018-dont}.\footnote{\url{https://github.com/EdinburghNLP/XSum}} For TransAbs, BertAbs, and BertExtAbs, we also directly use the model outputs released by \citet{liu2019text}.\footnote{\url{https://github.com/nlpyang/PreSumm}} For BART \cite{lewis2020bart} and PEGASUS \cite{zhang2020pegasus}, we take advantage of the XSum pretrained models released on HuggingFace\footnote{\url{https://huggingface.co/facebook/bart-large-xsum}, \url{https://huggingface.co/google/pegasus-xsum}} and generate summaries from them. Lastly, we finetune T5 large on XSum training set via Transformers of HuggingFace \cite{wolf-etal-2020-transformers} and generate summaries from the finetuned model. Table~\ref{table:pyrxsum} lists the ROUGE-2 (R2) \cite{lin2004rouge} results of the 10 systems evaluated only on the 100 examples. 

\begin{table*}[t]
\centering
\small
\resizebox{0.99\textwidth}{!}{%
\begin{tabular*}{1.04\textwidth}{l cc cc cc cc cc cc cc cc}
\toprule
& \multicolumn{8}{c}{System-level} & \multicolumn{8}{c}{Summary-level} \\
\cmidrule(lr){2-9} \cmidrule(lr){10-17}
& \multicolumn{2}{c}{TAC08} & \multicolumn{2}{c}{ TAC09} & \multicolumn{2}{c}{REALSumm} & \multicolumn{2}{c}{PyrXSum} & \multicolumn{2}{c}{TAC08} & \multicolumn{2}{c}{ TAC09} & \multicolumn{2}{c}{REALSumm} & \multicolumn{2}{c}{PyrXSum} \\
Metrics & $r$ & $\rho$ & $r$ & $\rho$ & $r$ & $\rho$ & $r$ & $\rho$ & $r$ & $\rho$ & $r$ & $\rho$ & $r$ & $\rho$ & $r$ & $\rho$ \\
\midrule
ROUGE-1 & .92 & .92 & .93 & .88 & .76 & .66 & \underline{\bf1.0} & \underline{\bf1.0} & .61 & .59 & .66  & .60 & .47 & .43 & .30 & .30\\
ROUGE-2 & .96 & \underline{\bf.96} & .97 & .93 & .82 & .82 & \underline{\bf1.0} & \underline{\bf1.0} & .63 & .60 & .67 & .61 & .43 & .41 & .30 & .30 \\
ROUGE-L & .93 & .91 & .95 & .90 & .82 & \underline{\bf.84} & \underline{\bf1.0} & \underline{\bf1.0} & .56 & .53 & .62 & .57 & .40 & .36 & .29 & .29 \\
AutoSummENG & .94 & .89 & .96 & .91 & .51 & .40 & .60 & .60 & .64 & .62 & .68 & .62 & .33 & .32 & .07 & .07 \\
METEOR & .94 & .93 & .95 & .88 & .81 & .76 & .60 & .60 & .65 & .62 & .70 & .65 & \underline{.49} & .46 & .26 & .26 \\
BEwT-E & .96 & .94 & .98 & .93 & .82 & .72 & .60 & .60 & .65 & .62 & .72 & .66 & .43 & .40 & .28 & .28 \\
S3 pyr & .95 & .93 & .95 & .89 & .81 & .78 & \underline{\bf1.0} & \underline{\bf1.0} & .66 & .63 & .71 & .65 & \underline{.49} & .45 & .24 & .24 \\
S3 resp & .96 & .94 & .96 & .90 & .82 & .82 & \underline{\bf1.0} & \underline{\bf1.0} & .66 & .64 & .71 & .65 & .48 & .44 & .22 & .22 \\
PyrEval qual & .91 & .88 & .91 & .84 & - & - & - & - & .40 & .38 & .46 & .42 & - & - & - & -\\
PyrEval comp & .90 & .87 & .93 & .80 & - & - & - & - & .41 & .39 & .49 & .44 & - & - & - & -\\
BertScore & .91 & .89 & .98 & .89 & .69 & .68 & .60 & .60 & .61 & .58 & .67 & .62 & .43 & .40 & .12 & .12 \\
BertScore (idf) & .93 & .90 & .97 & .89 & .70 & .68 & .60 & .60  & .61 & .58 & .68 & .63 & .44 & .41 & .10 & .10 \\
MoverScore & .95 & .92 & .96 & .90 & .47 & .46 & .20 & .20 & .64 & .61 & .71 & .65 & .37 & .34 & .14 & .14 \\
QAEval EM & .94 & .90 & .97 & .92 & .83 & .70 & .60 & .60 & .48 & .47 & .58 & .53 & .22 & .20 & .46 & .46 \\
QAEval F1 & .97 & .93 & .98 & .95 & \underline{\bf.86} & .78 & .20 & .20 & .61 & .58 & .66 & .60 & .31 & .29 & .42 & .42 \\
\midrule
Lite$^3$Pyramid & \underline{.98}& .95& \underline{\bf.99} & \underline{\bf.97} & .78 & .76 & .20 & .20 & \underline{.74}&.\underline{71} & \underline{.78}& \underline{.73} & \underline{.49} & \underline{.47} & \underline{.48$^*$} & \underline{.48$^*$} \vspace{4pt} \\
Lite$^{2.5}$Pyramid  & \bf.99 & \bf .96 & \bf.99& \bf.97 & .71 & .70 & .60 & .60 &.84 & .81& .86& .82 & .53 & .51 & .53$^*$ & .53$^*$\\
Lite$^2$Pyramid & .99  & .98  & \bf .99 & .98 & .74 & .72 & \bf1.0 & \bf1.0 & \bf .87 & \bf .84 & \bf .88 & \bf .84 & \bf.56 & \bf.52 & \bf.66$^*$ & \bf.66$^*$ \\
Lite$^2$Pyramid-0 & .88 & .85 & .97 & .90 & .73 & .72 & \bf1.0 & \bf1.0 & .62 & .60 & .71 & .66 & .48 & .47 & .63 & .63 \\
\bottomrule
 \end{tabular*}
 }
 \vspace{-5pt}
\caption{5-fold (split by systems) cross-validation results. In each column, the \textbf{bold} numbers are the best and the \underline{underline} numbers are the best out of automatic metrics.  All Lite$^2$Pyramid-0 numbers are based on $f_{\textsc{nli}}=l^{3c}$. All other numbers of our metrics are based on $f_{\textsc{nli}}=p^{2c}$, except that those star$^*$ numbers are based on $f_{\textsc{nli}}=l^{2c}$. }
\label{table:result_system}
\vspace{-7pt}
\end{table*}

Then, we collect the SCUs' presence labels for each system summary on Amazon Mechanical Turk. Figure~\ref{fig:pyrxsum} illustrates the data annotation instructions and interfaces shown to crowdsourcing workers. The summaries usually only contain one sentence. We estimate it will take around 30-45 seconds for a native English speaker to finish one HIT. Following \citet{bhandari-etal-2020-evaluating}, we pay \$0.15 per HIT, which is respectably higher than the U.S. federal minimum wage requirement. Meanwhile, we select annotators that are located in the U.S., have an approval rate greater than 98\%, and have at least 10,000 approved HITs. 

We collect 4 responses per summary (100 * 10 * 4 HITs) and finally, 104 workers were involved. After annotation, we filter the annotations from a noisy worker who did 210 HITs but disagreed with the majority in 72\% of the time. After
this filtering, we obtain an average inter-annotator agreement (Krippendorff’s alpha \cite{krippendorff2011computing}) of 0.73. Following \citet{bhandari-etal-2020-evaluating}, we use the majority vote to mark the presence of an SCU and break ties by “not present”. Table~\ref{table:pyrxsum} shows the gold Pyramid scores of different systems.

Usually judging the presence of SCUs is considered as a task with little ambiguity, reflected by the high inter-annotator agreements achieved by REAMSumm (0.66) \cite{bhandari-etal-2020-evaluating} and our PyrXSum (0.73). To further verify this, on REALSumm, instead of taking the majority vote, we randomly sample 1 out of 4 as the gold label. We conduct this for 3 rounds and test Lite$^2$Pyramid’s correlations with these 3 sets of human labels. We get 0.89/0.63, 0.90/0.63, 0.90/0.63 system/summary-level Pearson correlations, respectively. They are close to each other and also close to the results obtained from the majority vote (0.89/0.64). This means workers give rather consistent SCU-presence labels.

\subsection{Experimental Details}
\label{appendix:expdetails}
\paragraph{NLI.} For the natural language inference (NLI) used in our work, we take advantage of the pre-trained NLI released by \citet{nie-etal-2020-adversarial}.\footnote{\url{https://github.com/facebookresearch/anli}} We use the RoBERTa \cite{liu2019roberta} large based version.\footnote{ynie/roberta-large-snli\_mnli\_fever\_anli\_R1\_R2\_R3-nli} This model is implemented on HuggingFace's Transformers \cite{wolf-etal-2020-transformers} and was trained on SNLI \cite{bowman-etal-2015-large}, MNLI \cite{williams2018broad}, FEVER \cite{thorne-etal-2018-fever}, and ANLI \cite{nie-etal-2020-adversarial}. We directly use this pretrained model for our Lite$^2$Pyramid-0 metric. When we finetune this model, for simplicity, we always use learning rate=1e-5, linear schedule with warmup, and AdamW \cite{loshchilov2018decoupled} optimizer, and we always finetune for 2 epochs. 

\paragraph{SRL.} For Semantic Role Labeling (SRL) model, we use the out-of-the-box  SRL model pretrained by AllenNLP \cite{gardner2018allennlp}.\footnote{\url{https://demo.allennlp.org/semantic-role-labeling}} And it is based the model proposed by \citet{shi2019simple}.

\paragraph{Coreference Resolution.} For the Coreference Resolution model, we also use the out-of-the-box Coreference model pretrained by AllenNLP \cite{gardner2018allennlp}.\footnote{\url{https://demo.allennlp.org/coreference-resolution}} And it is based the model proposed by \citet{lee-etal-2018-higher}.

\paragraph{Constituency Parsing.} For the Constituency Parsing model, we also use the out-of-the-box parser pretrained by AllenNLP \cite{gardner2018allennlp}.\footnote{\url{https://demo.allennlp.org/constituency-parsing}} And it is based the model proposed by \citet{joshi-etal-2018-extending}.

\paragraph{Regressor.} The full features we used to train the regressor are: (1) sentence length (in words); (2) linearized parsing tree length (in characters); (3) parsing tree depth; (4) parsing tree depth divided by sentence length; (5) the counts of parsing tree non-terminal tokens.\footnote{WRB, RBR, ADVP, VBG, \$, '', WHADVP, -RRB-, JJR, NAC, PRP, NNS, WP, VBZ, MD, WDT, NP, ADJP, PDT, EX, UH, NN, NFP, SYM, PRP\$, RBS, FRAG, NX, CONJP, RP, WHPP, CC, VBD, LS, ., SBAR, TO, JJ, IN, VP, -LRB-, S, QP, SQ, CD, ``, X, POS, XX, PP, PRT, JJS, HYPH, ,, RB, VBN, :, VBP, DT, VB, SINV, UCP, WHNP, NNPS, NNP.} Then, we train the regressor through the XGBoost Python Package\footnote{\url{https://xgboost.readthedocs.io/en/latest/python/index.html}} and we set the max depth=3, learning rate eta=0.1, number of round=40. 

\subsection{Additional Results \& Ablations}

\begin{figure}
    \centering
    \includegraphics[width=0.47\textwidth]{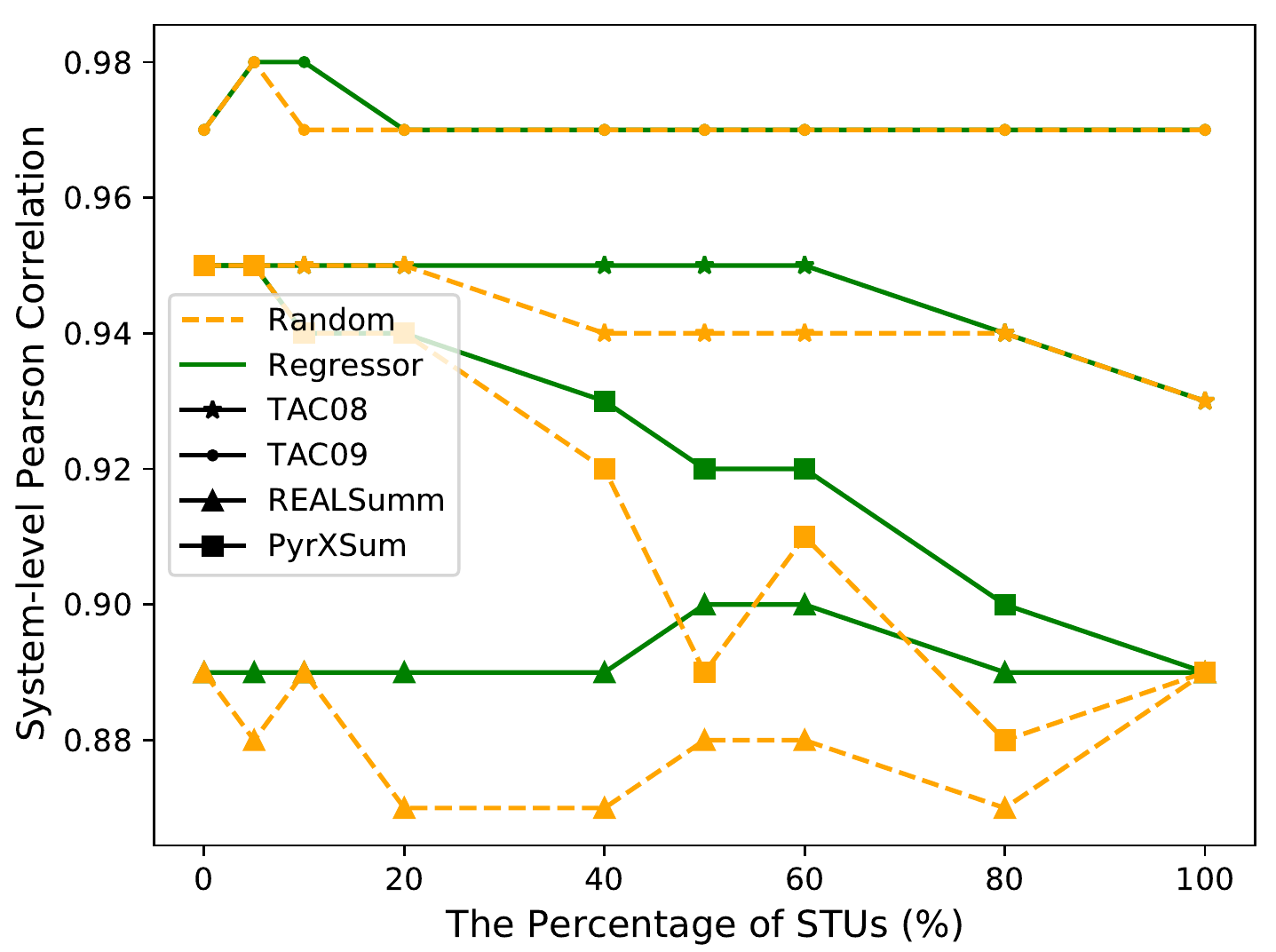}
    \vspace{-7pt}
    \caption{Lite$^{2.x}$Pyramid curves (for system-level correlations) and its comparison to replacing \emph{random} sentences' SCUs with STUs. }
    \vspace{-12pt}
    \label{fig:lite2x_sys}
\end{figure}

\paragraph{Cross-Validation Results.} As a complement of Figure 2 in the main paper, Figure~\ref{fig:lite2x_sys} shows the Lite$^{2.x}$Pyramid curves for system-level correlations. It can be observed that comparing to using random replacement, our Lite$^{2.x}$Pyramid always achieves higher or the same correlations when the same amount of human effort is reduced. Besides, Table~\ref{table:result_system} shows our 5-fold cross-validation (split \emph{by systems}) results.

\paragraph{NLI.} On REALSumm, the finetuned and non-finetuned NLI models get 82.34\% and 80.51\% accuracy for SCU-presence prediction, respectively. Similarly, 92.53\%/87.63\% are for TAC08, 93.25\%/88.66\% are for TAC09, and 92.45\%/91.13\% are for PyrXSum. Each number is an average of 5 folds (split by examples). As shown in Table~\ref{table:result_example}, Lite$^2$Pyramid (with finetuned NLIs) always gets higher correlations than Lite$^2$Pyramid-0 (with non-finetuned NLIs) except for PyrXSum. Therefore, we think NLI accuracy positively affects the results. In our work, we use a RoBERTa \cite{liu2019roberta} based NLI models. Here, to evaluate our metrics' robustness to different types of NLI models, we test an ALBERT \cite{lan2019albert} based NLI model.\footnote{ynie/albert-xxlarge-v2-snli\_mnli\_fever\_anli\_R1\_R2\_R3-nli} On REALSumm, Lite$^2$Pyramid gets 0.90/0.64 system/summary-level Pearson correlations with human, similar to our RoBERTa-NLI based results (0.89/0.64).

\paragraph{Regressor.} On REALSumm, TAC08, TAC09, and XSum, our regressors’ Mean Absolute Errors (MAE) are 0.135, 0.211, 0.206, and 0.090, respectively. On REALSumm, we test a weaker regressor (MAE=0.167), while we get similar results (0.89/0.62 system/summary-level Pearson correlations for Lite$^{2.5}$Pyramid) to our original regressor (0.90/0.62). However, the sentence selector guided by our regressor always works better than the random selector (shown in Figure~\ref{fig:lite2x} and Figure~\ref{fig:lite2x_sys}). We think the regressor influences the results by determining the ranking. If we reverse the ranking from the regressor, i.e., replacing SCUs with STUs for more complex sentences, we get lower correlations (0.88/0.60). In our work, we use XGBoost regressor instead of regressors based on pretrained LM because we think to determine the simulation easiness of sentences, syntactic features are more important than semantic features, and we want to keep the regressor as light-weight as possible. Here, we evaluate a RoBERTa-based regressor on REALSumm and it gets 0.89/0.62 system/summary-level Pearson correlations for Lite$^{2.5}$Pyramid, which is similar to our XGBoost regressor’s results (0.90/0.62).

\end{document}